\documentclass[10pt,technote,compsoc]{IEEEtran}

\ifCLASSOPTIONcompsoc
  \usepackage[nocompress]{cite}
\else
  \usepackage{cite}
\fi

\usepackage{graphicx}
\usepackage{amsmath,amssymb}
\usepackage{color}

\usepackage{microtype}
\usepackage{tabu}
\usepackage{bm}
\usepackage[labelfont=bf]{caption}
\usepackage{subcaption}
\usepackage[usenames,dvipsnames]{xcolor}
\usepackage[hidelinks]{hyperref}

\begin{document}

\title{Seeing with Humans: Gaze-Assisted\\Neural Image Captioning}

\author{Yusuke~Sugano
        and~Andreas~Bulling
\IEEEcompsocitemizethanks{
\IEEEcompsocthanksitem Y. Sugano and A. Bulling are with the Perceptual User Interfaces Group, Max Planck Institute for Informatics, Saarbr\"{u}cken, Germany.\protect\\
E-mail: \{sugano, bulling\}@mpi-inf.mpg.de}
}

\IEEEtitleabstractindextext{%
\begin{abstract}
%
Gaze reflects how humans process visual scenes and is therefore increasingly used in computer vision systems. Previous works demonstrated the potential of gaze for object-centric tasks, such as object localization and recognition, but it remains unclear if gaze can also be beneficial for scene-centric tasks, such as image captioning. We present a new perspective on gaze-assisted image captioning by studying the interplay between human gaze and the attention mechanism of deep neural networks. Using a public large-scale gaze dataset, we first assess the relationship between state-of-the-art object and scene recognition models, bottom-up visual saliency, and human gaze. We then propose a novel split attention model for image captioning. Our model integrates human gaze information into an attention-based long short-term memory architecture, and allows the algorithm to allocate attention selectively to both fixated and non-fixated image regions. Through evaluation on the COCO/SALICON datasets we show that our method improves image captioning performance and that gaze can complement machine attention for semantic scene understanding tasks.

\end{abstract}

\begin{IEEEkeywords}
Eye tracking, Attention, Image Captioning
\end{IEEEkeywords}
}

\maketitle

\IEEEdisplaynontitleabstractindextext


\section{Introduction}\label{sec:introduction}

Human gaze reflects processes of cognition and perception and therefore represents a rich source of information about the observer of a visual scene.
Consequently, gaze has successfully been used for tasks such as eye-based user modeling~\cite{bulling11_pami,fathi2012learning,papadopoulos2014training,sattar15_cvpr,karthikeyan2015eye} and opened up new opportunities to further advance human-machine collaboration -- collaborative human-machine vision systems in which part of the processing is carried out by the machine while another part is performed by a human and conveyed to the machine via gaze information.

As eye tracking techniques mature and become integrated into daily-life devices such as smart glasses, it is becoming more realistic to assume the availability of additional gaze information.
Recent advances in crowd-sourcing techniques~\cite{jiang2015salicon,rudoy2012crowdsourcing,xu2015turkergaze} and appearance-based estimation methods~\cite{funes2014geometric,sugano2014learning,xucong15cvpr} are also further paving the way for low-cost and large-scale gaze data collection.
For these reasons, gaze-enabled computer vision methods have been attracting increasing interest in recent years.
Prior work has typically focused on object-centric tasks, such as object localization~\cite{karthikeyan2015eye,5989830} or recognition~\cite{papadopoulos2014training,karthikeyan2013and}.
Although gaze information is potentially even richer for tasks that require holistic scene understanding~\cite{yun2013exploring,zelinsky2013understanding}, the integration of gaze information for scene-centric computer vision algorithms, such as for image captioning, has not yet been explored.

At the same time, attention mechanisms -- which are inspired by how humans selectively attend to visual input -- have recently become popular in machine learning literature~\cite{larochelle2010learning,mnih2014recurrent}.
Models that include attention mechanisms were shown to improve performance and efficiency for a variety of computer vision tasks, such as facial expression recognition~\cite{zheng2014neural} or image captioning~\cite{xu2015show}.
However, given that most current attention models are trained without any supervision from human attention, it remains unclear whether gaze information can further improve model performance.

This gap between human gaze and the machine learning algorithms' own feature localization capability is a fundamental question for the role of human gaze in computer vision tasks.
There has been remarkable progress in learning-based visual saliency models that demonstrated good performance for predicting where humans look in a purely bottom-up manner~\cite{borji2013state}.
Also, state-of-the-art convolutional neural nets (CNNs) were reported to be able to efficiently localize key points in visual scenes~\cite{long2014convnets}, indicating that machines can be better at task-specific feature localization than humans in some scenarios.
It remains unclear, however, whether human gaze can complement bottom-up visual information for feature localization.

\begin{figure}[t]
    \centering
    \includegraphics[width=\linewidth]{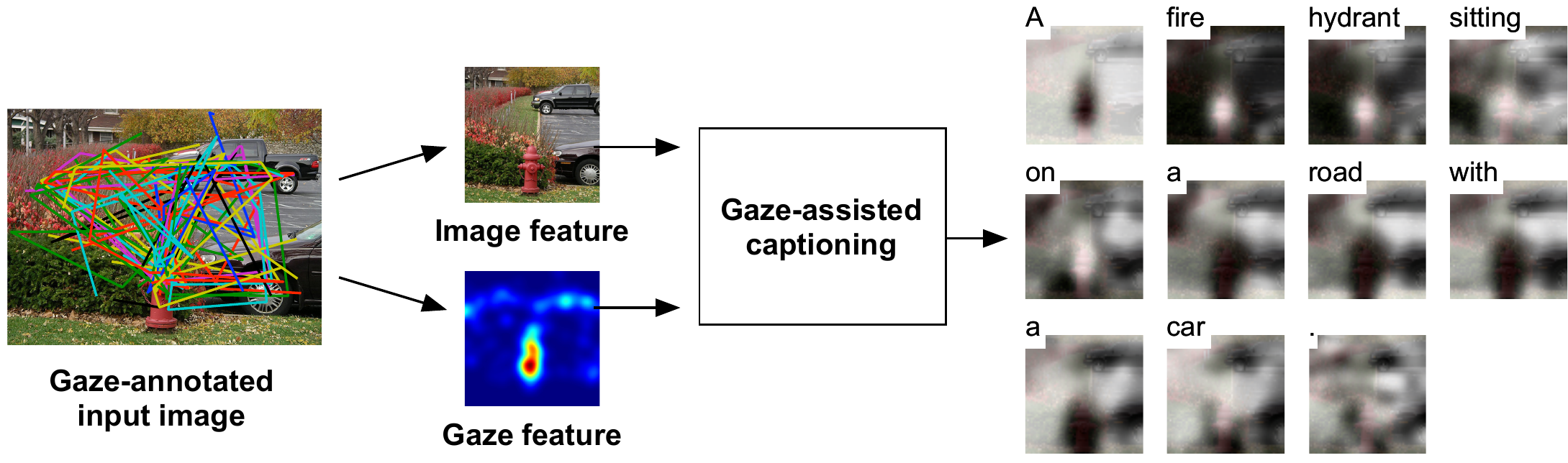}
    \caption{Our method takes gaze-annotated images as input, and uses both human gaze and bottom-up visual features for attention-based captioning.}
    \label{fig:concept}
\end{figure}

The goal of this work is to shed some light on these questions and to explore how the performance of image captioning models can be improved by incorporating gaze information.
Using the SALICON dataset~\cite{jiang2015salicon}, we first compare localization capability of state-of-the-art object and scene recognition models with human gaze.
Inspired by this analysis, we then propose a novel image captioning model (see \autoref{fig:concept}) that integrates gaze information into a state-of-the-art long short-term memory (LSTM) architecture with an attention mechanism~\cite{xu2015show}.
Human gaze is represented as a set of fixations, i.e., the static states of gaze upon a specific location, and the model localizes its machine attention selectively to both fixated and non-fixated regions.

The main contributions of this work are twofold.
First, we present an analysis of the relationship between object and scene recognition models and human gaze.
We take state-of-the-art CNN models as examples, and discuss how human gaze can influence their performance.
Second, we present a novel gaze-assisted attention mechanism for image captioning that is based on a split attention model.
We show that the proposed model improves captioning performance of the baseline attention model~\cite{xu2015show}.
To the best of our knowledge, this is the first work to 1) propose an actual model for gaze-assisted image captioning, and 2) relate human gaze input to deep neural network models of object recognition, scene recognition, and image captioning.
In this manner, this work provides the first unified overview on gaze, machine attention, image captioning and deep object/scene recognition models.


\section{Related Work}

Our gaze-assisted neural image captioning method is related to previous works in the emerging domain of collaborative human-machine vision as well as to attention mechanisms in current deep learning methods.

\subsubsection*{Gaze for Object Localization and Recognition}

The most well-studied approach is to use gaze information as a cue to infer object locations.
These efforts are motivated by findings from vision research that showed that humans tend to fixate on important or salient objects in the visual scene~\cite{einhauser2008objects,nuthmann2010object}.
The correlation between fixations and objects is also becoming important in the context of saliency prediction~\cite{li2014secrets,xu2014predicting}.
There have been several methods proposed for robust gaze-guided object segmentation~\cite{5989830,ramanathan2010eye,sugano2013graph}, and based on a similar assumption fixation information is also used for localizing important objects in videos~\cite{toyama2012gaze,Damen201698,karthikeyan2015eye}.
While these works mainly focused on object localization and the recognition was assumed to be purely vision-based,
there have been several attempts to further use gaze information for object recognition.
Gaze-related features were used for object recognition~\cite{karthikeyan2013and}, or incorporated into the image-based object recognition pipeline~\cite{yun2013studying,papadopoulos2014training,shcherbatyi2015gazedpm}.
However, the tasks were still object-centric and recognition tasks beyond object bounding boxes were not explored.

\subsubsection*{Gaze for Scene Description}

The usefulness of gaze information for image recognition is not limited to object-centric tasks.
For example, fixation locations were shown to provide information about the visual scene that can be used for both first-person and third-person activity recognition~\cite{fathi2012learning,mathe2015actions}.
However, only a few previous works explored the link between gaze and scene descriptions in the context of computer vision.
Subramanian et~al. identified correlations between gaze and scene semantics and showed that saccadic eye movements between different object regions can be used to discover object-object relations~\cite{subramanian2011can}. 
Yun et~al. showed that such correlations even exist in free-viewing conditions, i.e.\ without any specific tasks, and between image descriptions provided by different observers~\cite{yun2013studying,yun2013exploring}.
Coco and Keller further provided a detailed analysis on what kinds of visual guidance mechanisms cause such correlations~\cite{coco2015integrating}, and they demonstrated that annotators' gaze information can be used to predict the given captions~\cite{coco2012scan}.
However, all of these works provided only basic analyses of the link between gaze, visual feature, and scene descriptions and did not integrate gaze information into image captioning algorithms.

\subsubsection*{Attention Mechanisms in Deep Learning}

An increasing number of works investigate the potential of attention mechanisms for deep neural networks for computer vision tasks.
Attention mechanisms were shown to reduce computation and to make the networks more robust to changes in input resolution~\cite{larochelle2010learning,mnih2014recurrent,ranzato2014learning}.
Even more importantly, attention mechanisms add localization capabilities to neural networks, resulting in improved performance for target localization and recognition tasks~\cite{ba2014multiple,gregor2015draw,zheng2014neural}.
Information localization capabilities are also important for tasks such as machine translation~\cite{bahdanau2014neural} or semantic description~\cite{cho2015describing}.
With image captioning having recently emerged as a core task for deep neural networks~\cite{devlin2015language,donahue2014long,fang2014captions,karpathy2014deep,mao2015learning,vinyals2014show,chen2015mind,ushiku2015common}, this is where a promising link with human attention arises.
The benefit of the attention mechanism for image captioning has been demonstrated by Xu et~al.~\cite{xu2015show}, and You et~al. proposed to use the attention mechanism in the semantic space~\cite{you2016image}.
Although the visual attention mechanism of humans is often referred to as the inspirational source for these methods, none of them related their attention mechanism to actual human gaze data.
Instead, the attention mechanisms were treated as pure machine optimization tasks and were neither designed nor evaluated to resemble human attentive behavior.


\section{Gaze for Object and Scene Recognition}
\label{sec:analysis}

We first conduct fundamental analyses on the relationship between gaze and image recognition models on the SALICON dataset~\cite{jiang2015salicon}.
Object/scene category classification is key for holistic image understanding and our goal is to quantify whether and how human gaze can help state-of-the-art classification models.
In addition, we also compare human gaze and bottom-up visual saliency using the boolean map saliency (BMS) algorithm~\cite{zhang2015exploiting}, which is one of the best-performing saliency models with a publicly available implementation.
Extending on prior work~\cite{subramanian2011can,yun2013studying}, we evaluate the localization capabilities of state-of-the-art object recognition models in comparison with human gaze and bottom-up saliency, and provide the first analysis of scene recognition models.
We take the 16-layer VGGNet~\cite{simonyan2014very} architecture as an example and discuss the correlation with human gaze for both object and scene recognition.
For object recognition, we use Simonyan~et~al.'s pre-trained model~\cite{simonyan2014very} on the ILSVRC-2012 dataset~\cite{russakovsky2014imagenet}.
Similarly, Wang~et~al.'s pre-trained model~\cite{wang2015places205} on the Places205 dataset~\cite{zhou2014learning} is used for scene recognition.

\subsection{Dataset}

\begin{figure*}[t]
\centering
\begin{subfigure}[b]{0.49\linewidth}
\includegraphics[width=\linewidth]{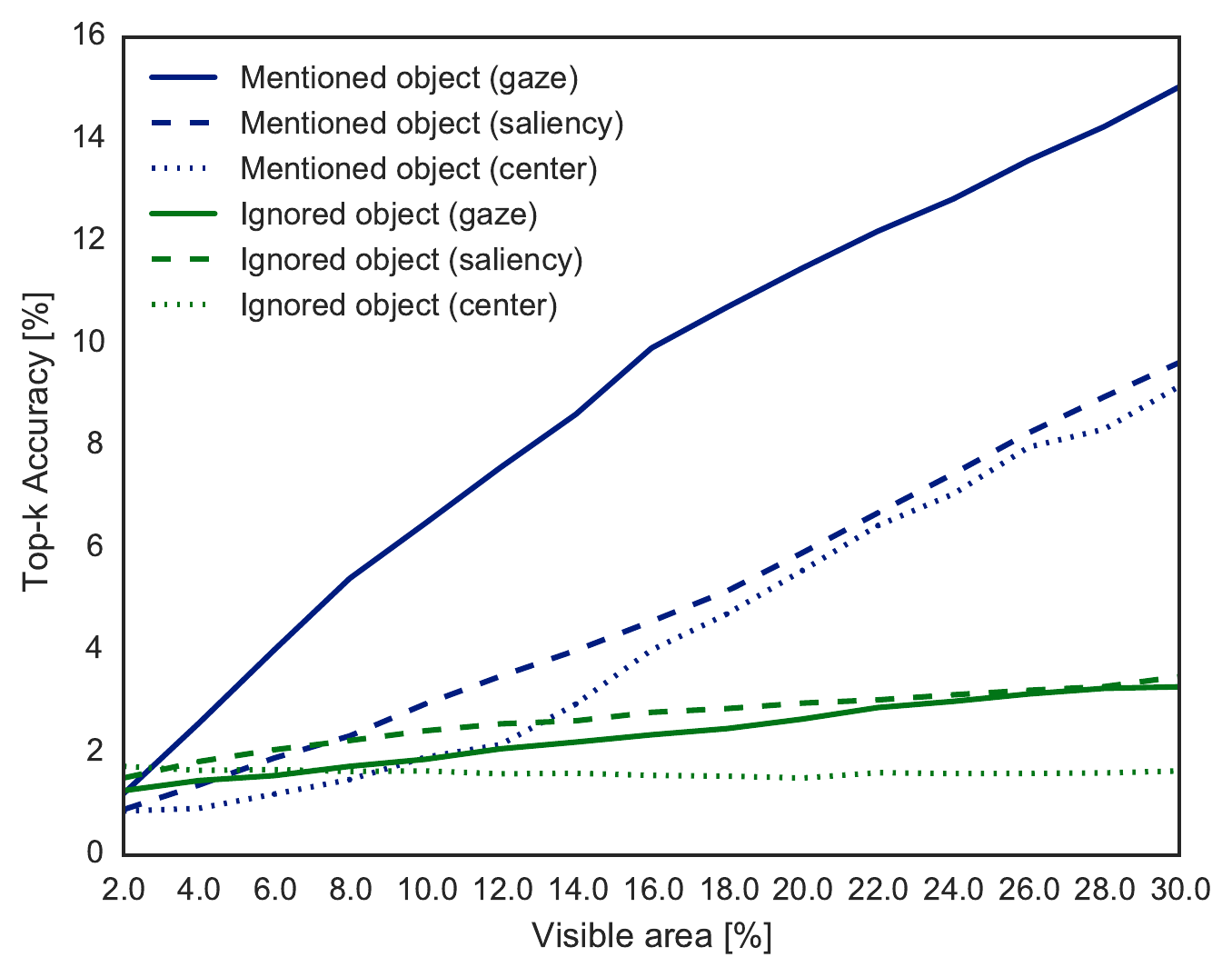}
\caption{Object recognition}
\label{fig:masked_accuracy_object}
\end{subfigure}
\begin{subfigure}[b]{0.49\linewidth}
\includegraphics[width=\linewidth]{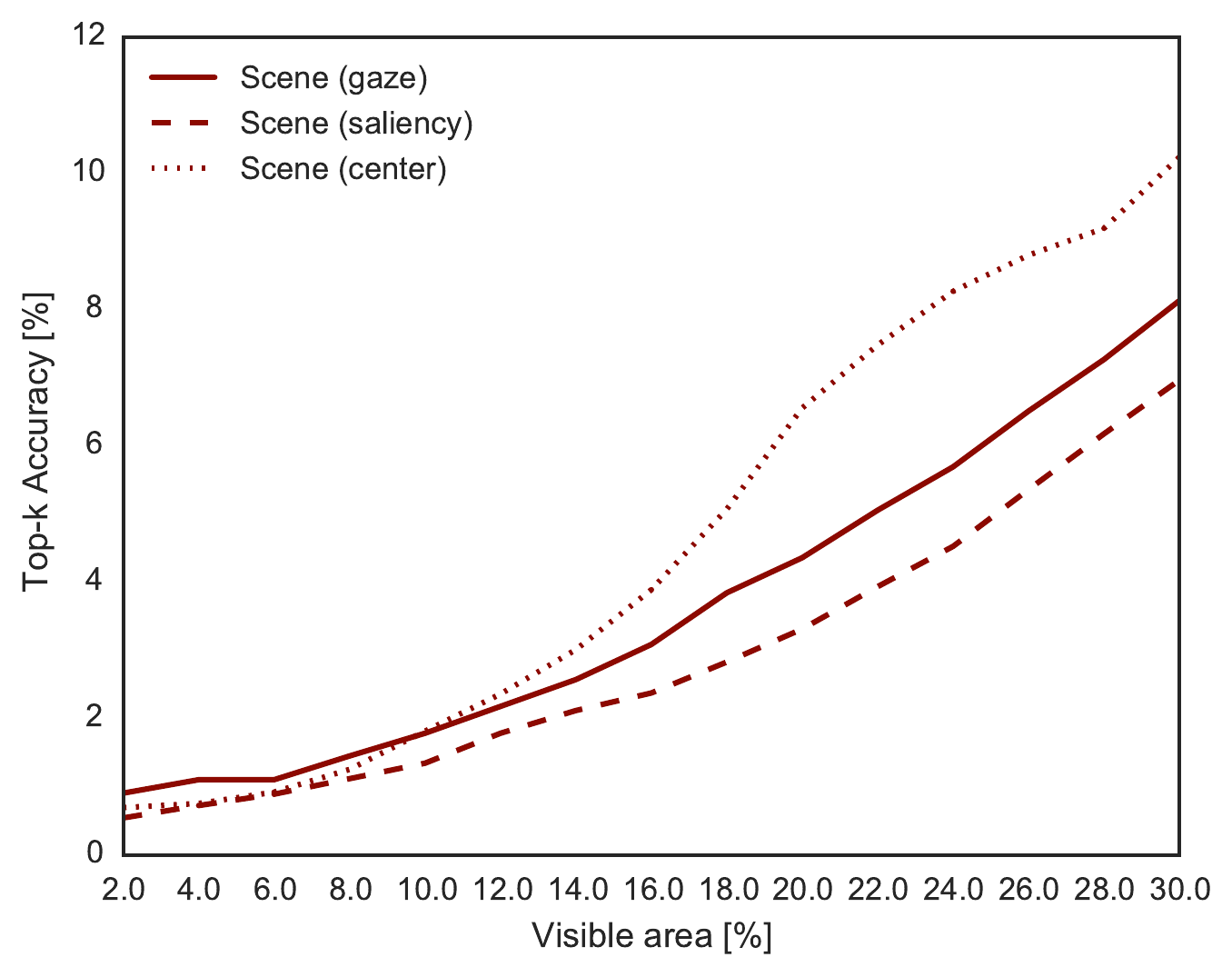}
\caption{Scene recognition}
\label{fig:masked_accuracy_scene}
\end{subfigure}
\caption{Top-$k$ accuracy for object and scene recognition. The horizontal axis indicates the ratio of the visible area given by thresholding the fixation, saliency and center maps. $k$ is set to the number of labels associated with each image.}
\label{fig:masked_accuracy}
\end{figure*}

The SALICON dataset provides gaze data collected through a crowd-sourced eye tracking experiment based on a mouse-contingent paradigm. 
The authors quantitatively showed that such mouse tracking data can resemble real eye tracking data under free viewing conditions.
The currently public part of the SALICON dataset contains 10,000 and 5,000 images from the training and validation sets of the COCO dataset, respectively.

Given that the SALICON dataset is a subset of the Microsoft COCO dataset~\cite{lin2014microsoft}, ground-truth gaze annotations can be obtained from the original annotations.
In addition, we assigned the original image annotations (captions, object segmentations, and categories) from the COCO dataset according to their image file names.
To associate the annotated $80$ object categories to the ILSVRC object categories, we used the WordNet synset IDs corresponding to the COCO object categories given by Chao et~al.~\cite{chao2015mining}.
We associated {\em all} object categories in the ILSVRC dataset that are children of the COCO categories in the WordNet hierarchy.

To study the relationship between image semantics and fixation locations in more detail, we further divided the object categories into two disjoint sets depending on whether they were mentioned in the captions.
We used the NLTK library~\cite{bird2009natural} to extract all nouns from the captions.
To reduce the effect of errors in the natural language processing, we only took nouns appearing more than twice in different captions.
On average, this process yielded $4.02$ nouns per image.
Then, all possible WordNet synsets corresponding to these nouns were extracted and associated with all child ILSVRC object categories in the same manner.
The object categories obtained from the COCO annotations were divided into two subsets, referred to as {\em mentioned} and {\em ignored} below, depending on whether the same synset was obtained from the captions.
While the {\em mentioned} subset corresponds to the ILSVRC object IDs found both in annotations and captions, the {\em ignored} subset contains IDs whose parent COCO ID had no correspondence in the caption ID set.
Given that the COCO dataset does not provide scene labels, we assigned scene categories according to the set of nouns extracted from the captions.
For some scene category names whose correspondences to WordNet synsets were ambiguous, we manually assigned IDs.
Finally, we extracted scene categories which have exactly the same synset in the set of nouns.

\subsection{Classification Performance with Fixation Masks}

\autoref{fig:masked_accuracy} shows the mean top-$k$ classification accuracy of object (both mentioned and ignored) and scene categories, when fixation/saliency maps were used.
Since in our setting each image can have multiple ground-truth object/scene labels, we set $k$ to the number of labels associated with each image.
The horizontal axis indicates the ratio of the visible area made by thresholding saliency/fixation maps, and the vertical axis shows classification scores where the model can only observe fixated or salient regions.
As a baseline, \autoref{fig:masked_accuracy} also shows an importance value plot with the same visible area ratio at the center of the input image.
We show performance values for both mentioned and ignored object categories independently.

In general, fixation and saliency information is more closely related to the object classification performance for the mentioned rather than ignored objects in \autoref{fig:masked_accuracy}.
It can also be seen that fixation maps achieve significantly better classification performance for mentioned objects than saliency maps.
This indicates that the performance gap between gaze and saliency is still significant in terms of recall.
It is in general more difficult for saliency models to suppress false positives, and the prediction result is not selective enough to support semantic image recognition.
If the ground-truth mentioned object segmentation is used for masking, the classification scores of mentioned and ignored objects become $24.50$\% and $3.54$\%, respectively.
Therefore, although the fixation mask cannot cover all of the important locations to discover all important objects, it can have roughly the same discrimination capability for non-important objects as ground-truth masks.
In contrast, for scene classification, the score is significantly lower at the beginning but becomes higher if the visible area is expanded.
While the score for fixation maps is consistently higher than for saliency maps, the center map baseline shows better performance than both fixation and saliency maps when the visible area is expanded to around $15$\%.

\begin{figure}[t]
\centering
   \includegraphics[width=0.95\linewidth]{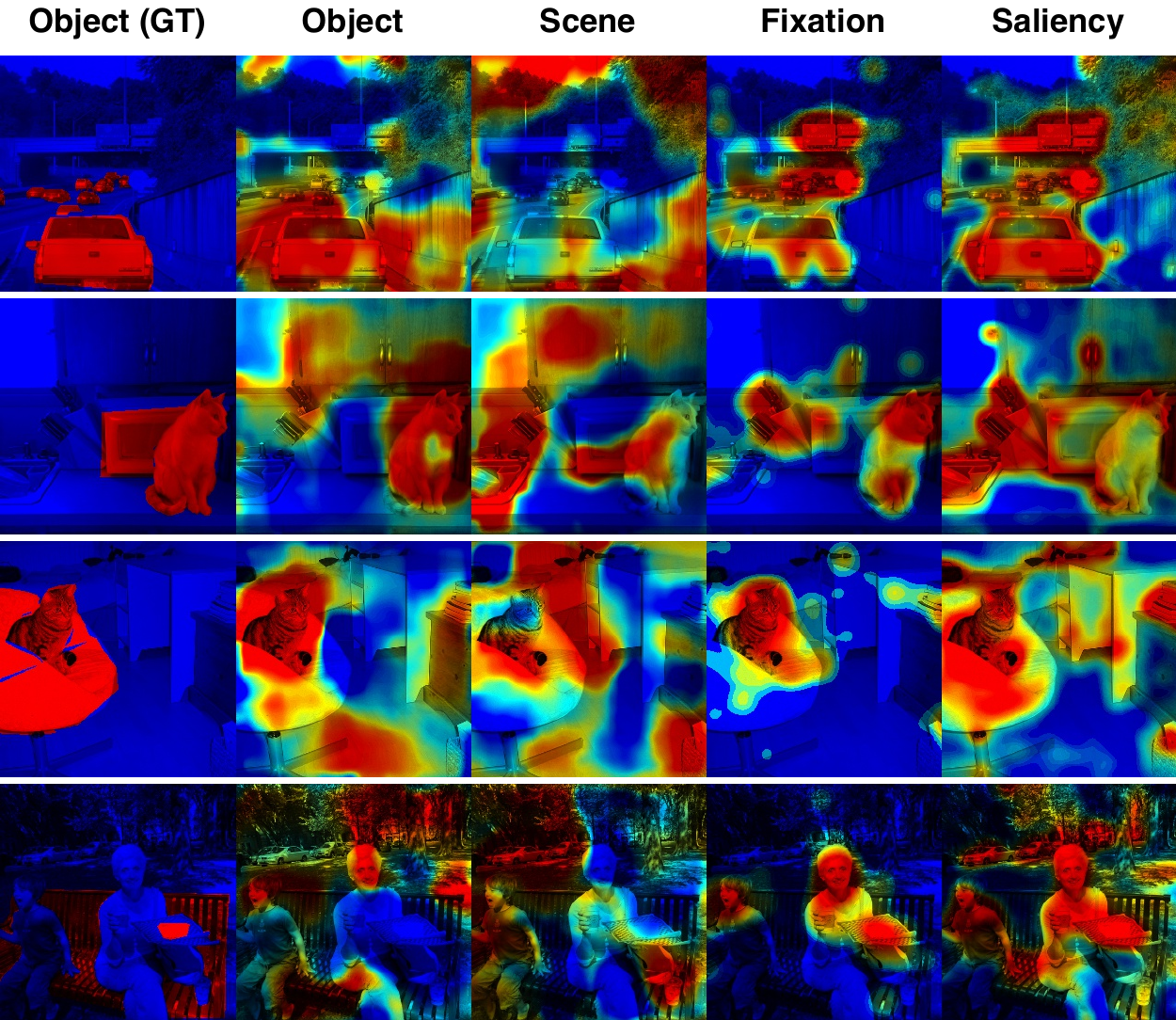}
\caption{Comparison of feature importance maps. Mean maps over all corresponding labels are overlaid onto the image with a color coding from blue (lowest importance) to red (highest importance).
For better visual comparison, all maps were histogram equalized.}
\label{fig:heatmap_samples}
\end{figure}

In addition, \autoref{fig:heatmap_samples} shows some examples of the feature importance maps for object and scene recognition models.
Following the approach of Zeiler et~al.~\cite{zeiler2014visualizing}, we evaluated the importance of image regions by measuring the decrease in recognition performance when the region is masked.
If the masked region contains important visual information for recognizing the target label, the recognition score should decrease substantially. Hence, the feature importance can be measured via a negative value of the score decrement.

The first column shows the ground-truth segmentations of mentioned objects obtained from the original COCO dataset annotation.
The second and third column show importance maps of the object and scene recognition models, respectively. 
The fourth column shows human fixation maps taken from the SALICON dataset, and the fifth column shows the corresponding purely bottom-up saliency prediction results.
While there is generally a larger similarity between fixation and object recognition maps than scene recognition maps, there are some object categories that do not attract human attention, such as the bench in the last image, even if the object is semantically important.

From these results, we can make several important observations.
First, fixation positions are indeed related to important locations for object recognition models to find semantically important objects.
By focusing on fixated regions, object recognition models can potentially discriminate between mentioned and ignored objects. 
The gap between human fixation and bottom-up saliency prediction is also related to this point, and human fixation gives better localization of important features.
Second, fixation positions are not significantly related to important locations for scene recognition models, and the area of focus has to be extended to find relevant information.


\section{Gaze-Assisted Image Captioning}
\label{sec:method}

\begin{figure*}[t]
\centering
   \includegraphics[width=\linewidth]{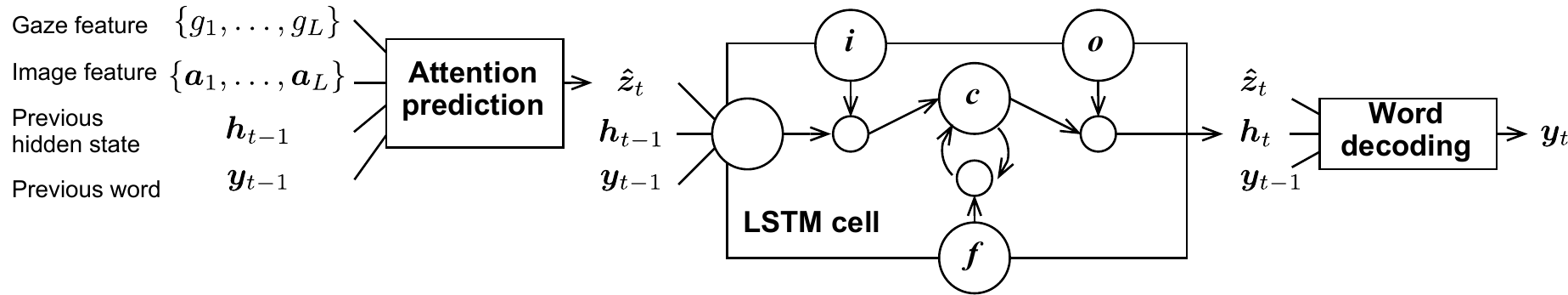}
\caption{Pipeline of the gaze-assisted image captioning. The attention function takes both image and gaze features as input, and the context vector weighted with the attention is given to the LSTM cell for word-by-word captioning.}
\label{fig:pipeline}
\end{figure*}

The previous analysis provides valuable insights into how gaze information can be exploited for semantic scene understanding and image captioning tasks.
While human gaze can help object recognition models to discriminate semantically important objects from non-important objects, the model also needs to pay attention to objects that do not attract human gaze.
Scene categories also cannot be fully recovered only from fixated regions even if they are important for semantic description.
Therefore, unlike humans, machines also need to obtain visual features from background regions to recognize scene categories.
Based on these observations, we propose to use gaze information to guide the attention-based captioning architecture~\cite{xu2015show} so that the model can allocate attention selectively according to fixation distribution.

We first briefly summarize Xu et~al.'s model based on the standard LSTM architecture~\cite{hochreiter1997long} and the soft attention mechanism~\cite{bahdanau2014neural}.
The input image is encoded as a set of $L$ feature vectors $a = \{\bm{a}_1, \dots, \bm{a}_L\}$ extracted from the last convolutional layer of an object recognition CNN.
Each $\bm{a}_i$ represents a $D$-dimensional feature vector corresponding to one part of the $L$ image regions.
The task is to output a caption $y = \{\bm{y}_1, \dots, \bm{y}_C\}$ encoded as a sequence of words.
The input to the LSTM cell is time step-dependent context vectors $\bm{\hat z}_t$ representing the specific part of the input image
\begin{eqnarray}
\bm{\hat z}_t &=& \sum_{i=1}^L \alpha_{t,i} \bm{a}_i,
\end{eqnarray}
which is defined as a weighted sum of the feature vectors.
The weight $\alpha_{t,i}$ represents the current state of the machine attention, and is defined as a function of the original image feature $\bm{a}_i$ and the previous hidden state $\bm{h}_{t-1}$ of the LSTM:
\begin{eqnarray}
e_{t,i} &=& f_{\textmd{att}} (\bm{a}_i, \bm{h}_{t-1}),\label{eq:attention}\\[0.25cm]
\alpha_{t,i} &=& \frac{\exp(e_{t,i})}{\sum_{i=1}^L \exp(e_{t,i})}.
\end{eqnarray}

Then, the caption sequence is estimated using the deep output layer as a function of the context vector, current hidden state, and the previous word~\cite{pascanu2013construct}:
$p(\bm{y}_t | \bm{a}, \bm{y}_1^{t-1}) = f_{\textmd{out}} (\bm{y}_{t-1}, \bm{h}_t, \bm{\hat z}_t)$.
They also introduced a doubly stochastic regularization $\sum_t \alpha_{t,i} \approx 1$ to the final cost function, i.e., the model is regularized so that attention is paid uniformly over the whole image.
Our interest in this work is to integrate human gaze feature into the attention function in Eq~\eqref{eq:attention}.

\subsection{Integration of Gaze Information}
\label{sec:model_fixation}

As illustrated in \autoref{fig:pipeline}, our method takes both image and gaze features as input.
The gaze feature is represented as a normalized fixation histogram $g = \{g_1, \dots, g_L\}$.
The fixation histogram obtained from gaze recording is cropped and resized to the same $L$ grid, and the $g_i$ is taken from the same grid as $\bm{a}_i$.
The gaze feature is given to the attention function together with the image feature $a$.
In the original model, $f_{\textmd{att}}$ is defined as a linear function
\begin{eqnarray}
e_{t,i} &=& \bm{w}_{\textmd{att}} \bm{p}_{t,i} + c_{\textmd{att}},\label{eq:attn_orig}
\end{eqnarray}
where $\bm{p}_{t,i}$ is a nonlinear projection of $\bm{a}_i$ and $\bm{h}_{t-1}$.
According to the previous analysis, the gaze-assisted attention function requires the flexibility to allocate attention to non-fixated regions.
Hence, we propose to split the machine attention according to human fixation as
\begin{eqnarray}
e_{t,i} = g_i \bm{w}_{\textmd{pos}} \bm{p}_{t,i} + (1-g_i) \bm{w}_{\textmd{neg}} \bm{p}_{t,i} + c_{\textmd{att}}.\label{eq:attn_gaze}
\end{eqnarray}
This model can learn different weights for fixated ($\bm{w}_{\textmd{pos}}$) and non-fixated ($\bm{w}_{\textmd{neg}}$) regions, and can efficiently utilize the gaze feature without losing too much information from non-fixated image regions.
We assess the advantage of this split attention model through experiments.

\subsection{Implementation Details}

Following the original implementation, the image features were extracted from the last convolutional layer before max pooling using the 19-layer VGGNet~\cite{simonyan2014very} without fine tuning.
This results in image features with $L = 14 \times 14 = 196$ and $D = 512$.
Input images were resized so that the shortest side had a length of 256 pixels, and the center cropped $224 \times 224$ image was given to the network.
Most of the other details were kept the same as in the original model implementation, while the dimensionality of the hidden state $\bm{h}$ was set to a lower value of 1,400 to account for the smaller amount of training data available.
In the experiments, all models were trained using the Adam algorithm~\cite{kingma2014adam}.


\section{Experiments}
\label{sec:experiments}

In this section, we report experimental results of our gaze-assisted image captioning models on the SALICON dataset.
Since the test set in the SALICON dataset does not provide gaze data, we randomly split the 5,000 validation images into two 2,500 images for validation and test.
The validation set is used for early stopping, and we show the performance of the best model on the test set.
As suggested by Xu et~al., we used the BLEU score~\cite{papineni2002bleu} on the validation set for model selection. 
In order to give a fair comparison given the limited amount of training data, we fixed the random seed value for weight initialization and mini-batch selection between different models. 
Hence, all models were trained under exactly the same conditions.

\subsection{Captioning Performance}

\begin{table*}[t]
\begin{center}
\caption{Image captioning performance. Columns show BLEU 1-4, METEOR, ROUGE$_L$ and CIDEr scores of different models. Columns show the performance of the original model~\cite{xu2015show}, the model without the attention term for non-fixated regions, the model using saliency maps instead of fixation maps, and the proposed split attention model. 
}
\label{tab:evaluation}
\begin{tabu}{X[l1.5]|X[c0.8]X[c0.8]X[c0.8]X[c0.8]|X[c1.2]|X[c1]|X[c0.9]}
         & \multicolumn{4}{c|}{BLEU} &      &      &        \\
Model    & 1  & 2  & 3  & 4  & METEOR & ROUGE & CIDEr \\ 
\hline
Machine~\cite{xu2015show} &   0.706   &   0.495   &   0.342   &   0.237   &   0.218   &   0.520   &   0.626     \\
Gaze-only &   0.704   &   0.492   &   0.340   &   0.236   &   0.215   &   0.519   &   0.613     \\
Saliency &   0.708   &   0.496   &   0.342   &   0.236   &   0.217   &   0.519   &   0.623     \\
Split attention &   {\bf 0.714}   &   {\bf 0.505}   &   {\bf 0.352}   &   {\bf 0.245}   &   {\bf 0.219}   &   {\bf 0.524}   &   {\bf 0.638}     \\
\end{tabu}
\end{center}
\end{table*}

We first show quantitative comparison of captioning performance.
As evaluation metrics we used the implementations of commonly used metrics (BLEU 1-4, METEOR, ROUGE$_L$ and CIDEr scores) provided with the COCO dataset~\cite{chen2015microsoft}.
BLEU is a standard machine translation score measuring co-occurrences of $n$-grams~\cite{papineni2002bleu}, and ROUGE$_L$ is a text summarization metric based on the longest common subsequence~\cite{lin2004rouge}.
METEOR score is based on the word alignment~\cite{denkowski2014meteor}, and CIDEr measures consensus in captions~\cite{vedantam2014cider}.
Although the limitations of these computational metrics have often been pointed out and it is fundamentally difficult to evaluate goodness of natural language captions, they give us some insights into how the additional gaze information changes the captioning performance.

In \autoref{tab:evaluation}, we report BLEU 1-4, METEOR, ROUGE$_L$ and CIDEr scores of different models.
In addition to the original machine attention model ({\em Machine})~\cite{xu2015show} as well as our gaze-assisted split attention model ({\em Split attention}), we also show two additional baseline models.
In the second row ({\em Gaze-only}), we show the case where the weight for non-fixated region in Eq.~(\ref{eq:attn_gaze}) is not used.
This model represents a more straightforward design of the gaze-assisted attention model which heavily relies on fixated regions.
In addition, we show results when the BMS saliency maps are used instead of fixation maps with the same architecture as the proposed model in the third row ({\em Saliency}).

As can be seen from the table, the proposed split attention model performs consistently better than the baseline models for all metrics even with the same underlying architecture.
The difference between {\em Gaze-only} and {\em Split attention} results clearly illustrates the significance of the proposed split attention architecture.
The performance gap between human gaze and bottom-up saliency also indicates the fundamental importance of the gaze information.
Together with the previous analysis, our results indicate the importance of human gaze information for semantic image understanding.

\subsection{Attention Allocation Examples}

\begin{figure*}[t]
\centering
\includegraphics[width=0.95\linewidth]{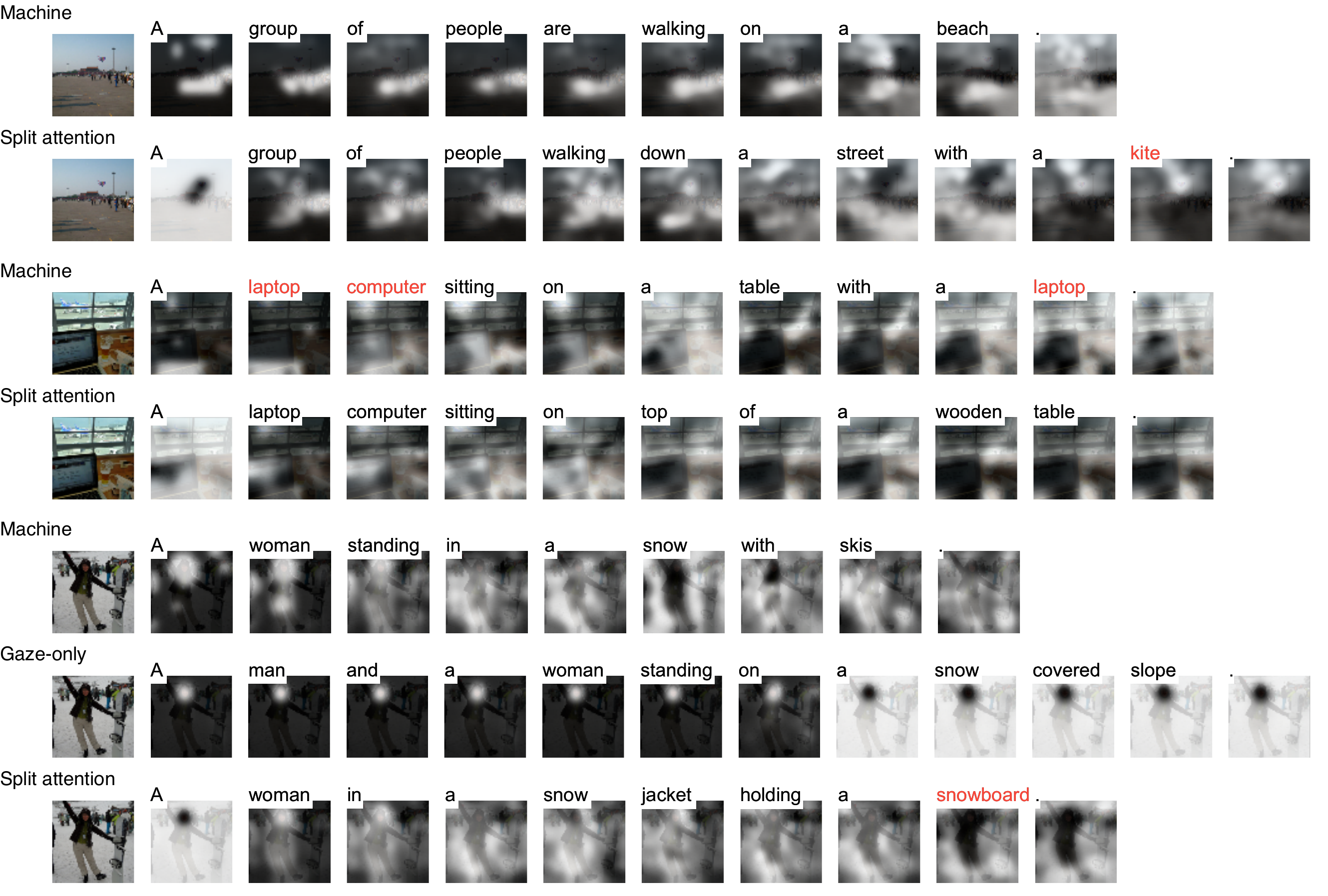}
\caption{Sample images, the machine attention map at each step as well as the corresponding output words for the baseline and gaze-assisted models. The first example illustrates the case where the proposed model finds small but important objects ({\em kite}) in the scene. It also helps to suppress the repetition of object description in cluttered scenes ({\em laptop} in the second example). The proposed split attention model can also describe objects without strong fixation, such as {\em snowboard} in the third example. See the supplementary material for more examples.
}
\label{fig:examples_fixation}
\end{figure*}

As discussed above, computational metrics do not fully explain how the captioning model is improved by the proposed attention model, and it is more important to discuss concrete examples of generated captions and their corresponding attention allocation results.
\autoref{fig:examples_fixation} visualizes some examples of the attention allocation results of the baseline and gaze-assisted models.
Each row shows the input image, attention map examples at each step $t$, and their corresponding output words.
Gaze information typically helps the model to find small but important objects in the scene, like the kites in the first example in \autoref{fig:examples_fixation}.
Since gaze information explicitly provides object locations, it is also beneficial for avoiding repetition of object discovery in cluttered scenes.
While the baseline model describes the laptop twice in the second example, the proposed model properly allocates the attention to the object region and generates the word only once.
It is also noteworthy that, although there are some objects which do not attract human fixations, the proposed split attention model has the flexibility to allocate attention to such objects.
In the third example, the {\em Gaze-only} model fails to describe the snowboard; the proposed split attention model successfully describes it.

Many prior works have reported the strong relationship between eye movements and viewing task~\cite{yarbus1967eye}, and how humans look at images heavily depends on the task the viewer is performing.
From these examples it can be seen that the behavior of the gaze-assisted model better resembles how humans see images, and they indicate the potential of utilizing different types of gaze behavior for task-oriented captioning.
In addition, the performance gap between human gaze and visual saliency poses another important research question; whether computational saliency models can achieve similar performance, especially with recently proposed deep models~\cite{liu2015predicting,huang2015salicon}.

\subsection{Word Prediction Performance}

\begin{figure*}[t]
\centering
\begin{subfigure}[b]{0.49\linewidth}
\includegraphics[width=\linewidth]{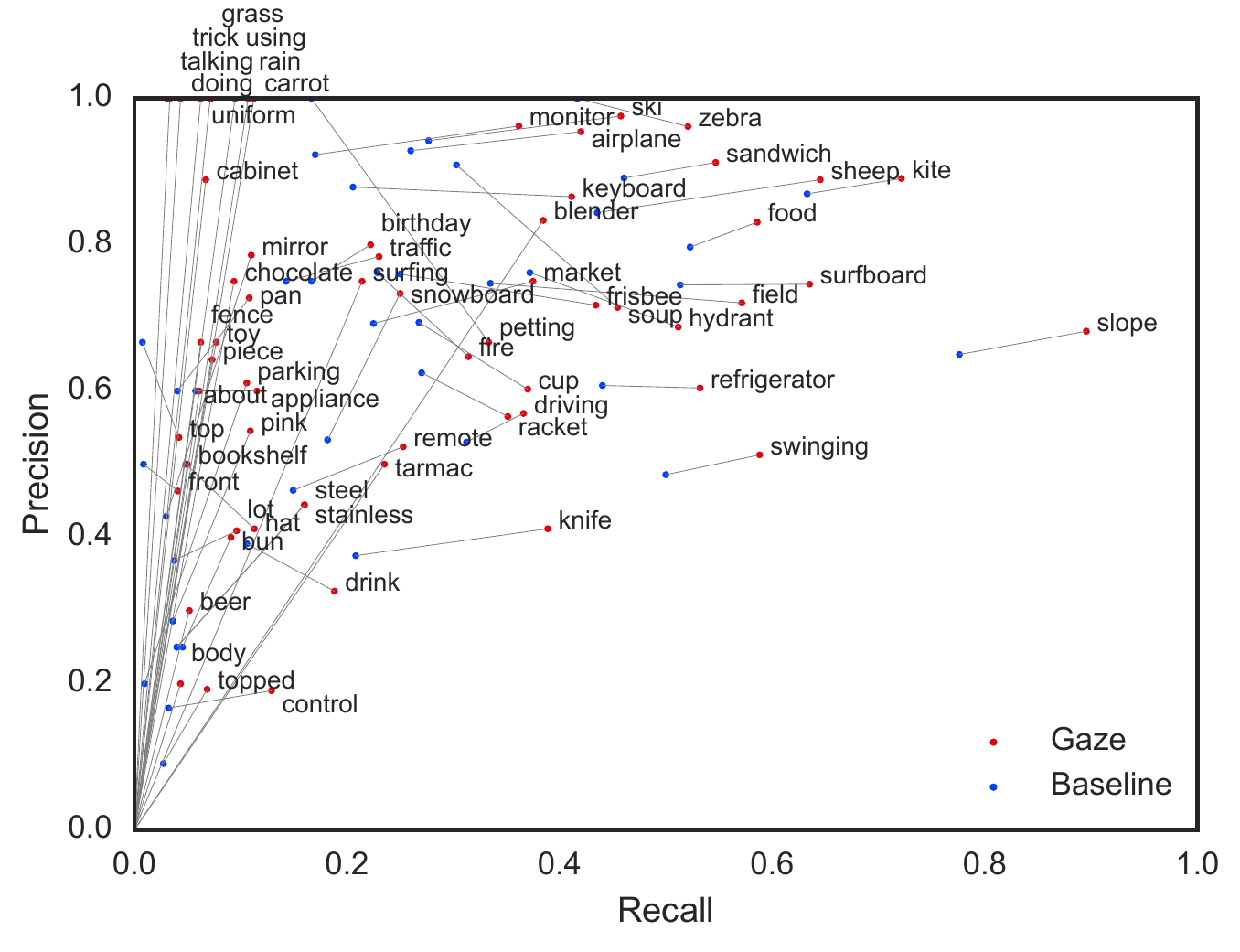}
\caption{}
\label{fig:word_gaze}
\end{subfigure}
\begin{subfigure}[b]{0.49\linewidth}
\includegraphics[width=\linewidth]{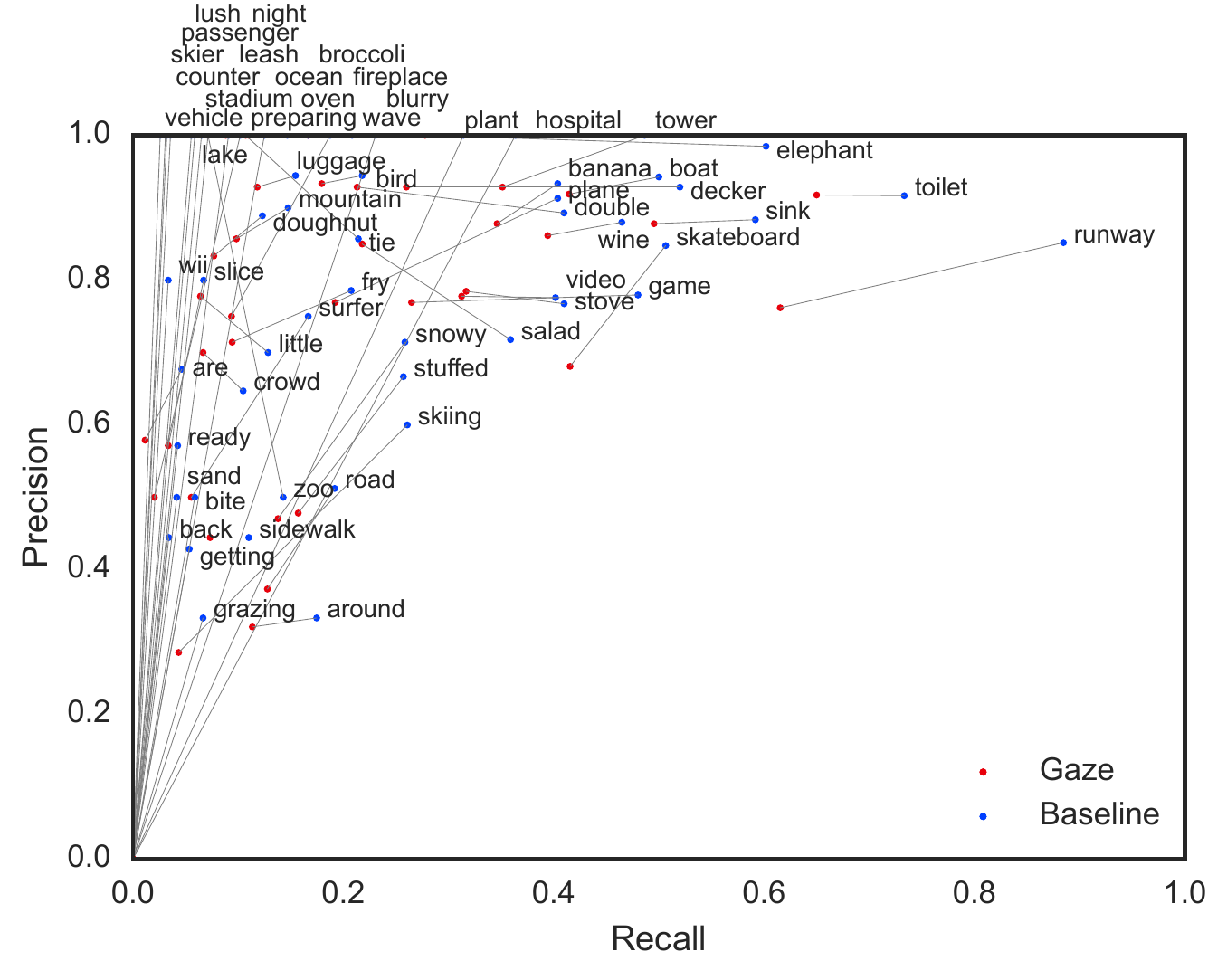}
\caption{}
\label{fig:word_base}
\end{subfigure}
\caption{Precision and recall of individual words for which the F-scores (a) improved and (b) degraded by using the proposed gaze-assisted image captioning model.}
\label{fig:word_pr}
\end{figure*}

To better understand which words are correctly discovered by the proposed gaze-assisted model, we analyzed individual word prediction scores.
In \autoref{fig:word_gaze}, we plot precision-recall points for words whose F-scores are improved by more than a threshold ($0.05$) by the fixation model.
To improve intelligibility, we show only words appearing more than $10$ times in the ground-truth captions.
We used the same weighted precision/recall measures as Chen et~al.~\cite{chen2015microsoft}; while negative images have a weight of $1$, the number of captions containing the word is used as a weight for positive images.
Words in both predicted and ground-truth captions are lemmatized using the NLTK library.
\autoref{fig:word_base} shows the same precision-recall plot for words whose recall scores become worse with the proposed model.

As discussed above, the proposed model improves word discovery scores for small important objects such as {\em kite}, {\em knife}, {\em umbrella}, {\em (fire) hydrant}, {\em traffic (sign)}.
The improvement of the words like {\em top}, {\em front}, and {\em about} could be related to the above-mentioned switching of the description subject.
Although the proposed attention function can attend to background regions, it still loses some performance on words related to background scene categories, such as {\em tower}, {\em table}, {\em runway}, and {\em fireplace}, as can be seen in \autoref{fig:word_base}.
It can also be seen that some words related to activity or context, like {\em game}, {\em blurry}, {\em night}, are also losing performance.
However, as illustrated in \autoref{fig:examples_fixation}, the proposed split attention model also helps the model to find objects which do not attract human attention, such as {\em snowboard} and {\em ski}.
Since the word discovery performance should also depend on the amount of training data, investigation of larger-scale eye tracking experiments is one of the most important future directions.

\section{Conclusion}

In this paper we presented a detailed study on how human gaze information can help holistic image understanding and captioning tasks.
We first analyzed the relationship between gaze and state-of-the-art recognition models for both object and scene categories.
We showed that human gaze is more correlated with important locations for object recognition models and can help to find more semantically important objects than bottom-up saliency models. 
We further presented the first gaze-assisted image captioning model and quantified its performance. 
With the previous findings in mind, we proposed a split attention model where the machine attention can be allocated selectively to fixated and non-fixated image regions.
Our model improves the captioning performance of a baseline model on the challenging COCO/SALICON dataset, and achieved a similar performance improvement compared to state-of-the-art bottom-up saliency models.

These results underline the potential of gaze-assisted image captioning, particularly for cluttered images without a clearly depicted central object. 
The approach is similarly appealing for gaze-assisted captioning of unorganized image streams, for example those recorded using life-logging devices or other egocentric cameras.
Since the SALICON dataset was collected using a pseudo-eye tracking setup, investigating gaze-assisted image understanding in an egocentric setting with real gaze data is one of the most interesting directions for future work.

\ifCLASSOPTIONcompsoc
  \section*{Acknowledgments}
\else
  \section*{Acknowledgment}
\fi

This work was funded, in part, by the Cluster of Excellence on Multimodal Computing and Interaction (MMCI) at Saarland University, as well as an Alexander von Humboldt Fellowship for Postdoctoral Researchers, and a JST CREST research grant.

\bibliographystyle{IEEEtran}
\bibliography{main}

\end{document}